\documentclass[11pt, a4paper, logo, twocolumn, copyright]{googledeepmind}

\usepackage[authoryear, sort&compress, round]{natbib}
\usepackage[textwidth=18mm]{todonotes}
\usepackage{graphicx}
\usepackage{makecell}
\usepackage{multirow}

\newcommand\blfootnote[1]{%
  \begingroup
  \renewcommand\thefootnote{}\footnote{#1}%
  \addtocounter{footnote}{-1}%
  \endgroup
}

\renewcommand{\today}{2024-06-27}

\title{Gemma 2: Improving Open Language Models at a Practical Size}

\author{Gemma Team, Google DeepMind\authfootnotemark{1}}

\authfootnotetext{1}{See Contributions and Acknowledgments section for full author list. Please send correspondence to {\tt gemma-2-report@google.com}.}

\begin{abstract}
In this work, we introduce Gemma 2, a new addition to the Gemma family of lightweight, state-of-the-art open models, ranging in scale from 2 billion to 27 billion parameters.
In this new version, we apply several known technical modifications to the Transformer architecture, such as interleaving local-global attentions~\citep{beltagy2020longformer} and group-query attention~\citep{ainslie2023gqa}.
We also train the 2B and 9B models with knowledge distillation~\citep{hinton2015distilling} instead of next token prediction.
The resulting models deliver the best performance for their size, and even offer competitive alternatives to models that are 2-3$\times$ bigger.
We release all our models to the community.
\end{abstract}

\begin{document}

\maketitle

\section{Introduction}

Large language models (LLMs) have demonstrated strong capabilities in language understanding, generation, and reasoning \citep{radfordlanguage,t5, gpt}. 
Scaling has been key to this recent progress, with many new capabilities only emerging at scale~\citep{gpt}.
The newest large models not only reach unprecedented performance on reasoning benchmarks~\citep{achiam2023gpt}, but they also demonstrate multimodal and multilingual capabilities~\citep{geminiteam2024gemini} and even the ability to use context lengths of over 1M tokens~\citep{geminiteam2024gemini}. 

Small-scale models have also shown a rapid increase in performance, but these gains are largely derived from increasing the length of training~\citep{llama1,mistral,gemmateam2024gemma}. 
This approach only scales logarithmically with dataset size~\citep{hoffmann2022training}, and the latest small models require up to 15T tokens to improve the state of the art by less than 1-2\%~\citep{llama3}.

Yet, these continued improvements provide evidence that small models are still under-trained.
In this work, we explore alternatives to improve small model performance without solely increasing training length. One solution is to improve the quality of information received by the network at each training step by replacing the next token prediction task with a richer objective. 

In particular, we focus our efforts on knowledge distillation~\citep{hinton2015distilling}, which replaces the one-hot vector seen at each token with the distribution of potential next tokens computed from a large model.
This approach is often used to reduce the training time of smaller models by giving them richer gradients.
In this work, we instead train for large quantities of tokens with distillation in order to simulate training beyond the number of available tokens.
Concretely, we use a large language model as a teacher to train small models, namely 2B and 9B models, on a quantity of tokens that is more than 50$\times$ the compute-optimal quantity predicted by the theory~\citep{hoffmann2022training}.
Along with the models trained with distillation, we also release a 27B model trained from scratch for this work.

\vspace{-.1cm}
We also leverage several known modifications of Transformers, namely the interleaving of global and local attention layers from~\cite{beltagy2020longformer}, and the Grouped-Query Attention~(GQA) mechanism of~\cite{ainslie2023gqa}.

\vspace{-.1cm}
Overall, Gemma 2 significantly advances state-of-the-art performance relative to comparable-scale open models and are even competitive with some models more than twice their size \citep{grok1, llama3, mistral, falcon}, across a variety of automated benchmarks and human evaluations. Example domains include question answering \citep{boolq, natural-questions}, commonsense reasoning \citep{winogrande, bbhard}, mathematics and science \citep{gsm8k, mmlu}, and coding \citep{mbpp, humaneval}.

While thorough testing of our models has been conducted, these tests cannot cover all applications and scenarios in which Gemma 2 may be used. With this in mind, all Gemma 2 users should conduct rigorous safety testing specific to their use case before deployment or use.

In this technical report, we provide an overview of models, including the architecture, training, and pre- and post-training recipes for Gemma 2. We also provide detailed evaluations across a wide variety of quantitative and qualitative benchmarks, as well as both standard academic benchmarks and human-preference evaluations. Finally, we discuss our approach to safe and responsible deployment and outline the broader implications of Gemma 2, its limitations, and advantages.

\begin{table}[t]
    \centering
    \setlength{\tabcolsep}{.4em}
    \begin{tabular}{l r r r}
    \toprule
        Parameters & \textbf{2B} & \textbf{9B} & \textbf{27B} \\
        \midrule
        \textit{d}\_{model} & 2304 & 3584 & 4608 \\
        Layers & 26 & 42 & 46  \\
        Pre-norm & yes & yes & yes \\
        Post-norm & yes & yes & yes \\
        \midrule
        Non-linearity & GeGLU & GeGLU & GeGLU \\
        Feedforward dim & 18432 & 28672 & 73728  \\
        \midrule
        Head type & GQA & GQA & GQA \\
        Num heads & 8 & 16 & 32 \\
        Num KV heads & 4 & 8 & 16  \\
        Head size & 256 & 256 & 128  \\
        Global att. span & 8192 & 8192 & 8192\\
        Sliding window & 4096 & 4096 & 4096\\
        \midrule
        Vocab size & 256128 & 256128 & 256128 \\
        Tied embedding & yes& yes & yes\\
    \bottomrule
    \end{tabular}
    \caption{Overview of the main model parameters and design choices. See the section on model architectures for more details.}
    \label{tab:model_params}
\end{table}

\section{Model Architecture}
\label{sec:architecture}

Similar to previous Gemma models \citep{gemmateam2024gemma}, the Gemma 2 models are based on a decoder-only transformer architecture~\citep{DBLP:journals/corr/VaswaniSPUJGKP17}.
We summarize the main parameters and architecture choices in Table \ref{tab:model_params}. 

A few architectural elements are similar to the first version of Gemma models; namely, a context length of 8192 tokens, the use of Rotary Position Embeddings (RoPE)~\citep{rope}, and the approximated GeGLU non-linearity~\citep{geglu}. A few elements differ between Gemma 1 and Gemma 2, including using deeper networks. We summarize the key differences below.

\noindent\textbf{Local Sliding Window and Global Attention}. We alternate between a local sliding window attention~\citep{slidingwindow, beltagy2020longformer} and global attention~\citep{globalattention} in every other layer. The sliding window size of local attention layers is set to 4096 tokens, while the span of the global attention layers is set to 8192 tokens.

\noindent\textbf{Logit soft-capping}. We cap logits~\citep{bello2016combinatorial} in each attention layer and the final layer such that the value of the logits stays between $-\text{soft\_cap}$ and $+\text{soft\_cap}$. More specifically, we cap the logits with the following function: $$\text{logits} \leftarrow \text{soft\_cap} * \tanh(\text{logits} / \text{soft\_cap}).$$
We set the soft\_cap parameter to $50.0$ for the self-attention layers and to $30.0$ for the final layer. 

\begin{table}[t]
    \centering
    \begin{tabular}{l r r}
    \toprule
        Model & \makecell{Embedding\\Parameters} & \makecell{Non-embedding\\Parameters} \\
        \midrule
        \textbf{2B} & 590,118,912 & 2,024,517,888 \\
        \textbf{9B} & 917,962,752 & 8,324,201,984 \\
        \textbf{27B} & 1,180,237,824 & 26,047,480,320 \\
    \bottomrule
    \end{tabular}
    \caption{Parameter counts for the Gemma models. 
    We inherit from the large Gemini vocabulary (256k entries), that is designed to work on a large number of languages, hence, the larger embedding parameter counts compared to models that are limited to one or a few languages.}
    \label{tab:model_param_counts}
\end{table}

\noindent\textbf{Post-norm and pre-norm with RMSNorm}. To stabilize training, we use RMSNorm~\citep{rmsnorm} to normalize the input and output of each transformer sub-layer, the attention layer, and the feedforward layer.

\noindent\textbf{Grouped-Query Attention}~\citep{ainslie2023gqa}. 
We use GQA with $\text{num\_groups} = 2$, based on ablations showing increased speed at inference time while maintaining downstream performance.

\section{Pre-training}

We provide a brief overview of the parts of our pre-training that differs from Gemma 1.

\subsection{Training Data}

We train Gemma 2 27B on 13 trillion tokens of primarily-English data, the 9B model on 8 trillion tokens, and the 2B on 2 trillion tokens. 
These tokens come from a variety of data sources, including web documents, code, and science articles. Our models are not multimodal and are not trained specifically for state-of-the-art multilingual capabilities.
The final data mixture was determined through ablations similar to the approach in Gemini 1.0~\citep{geminiteam2023gemini}.
 
\noindent\textbf{Tokenizer.}
We use the same tokenizer as Gemma 1 and Gemini: a SentencePiece tokenizer with split digits, preserved whitespace, and byte-level encodings \citep{kudo-richardson-2018-sentencepiece}.
The resulting vocabulary has 256k entries.

\noindent\textbf{Filtering.}
We use the same data filtering techniques as Gemma 1. Specifically, we filter the pre-training dataset to reduce the risk of unwanted or unsafe utterances, filter out certain personal information or other sensitive data, decontaminate evaluation sets from our pre-training data mixture, and reduce the risk of recitation by minimizing the proliferation of sensitive outputs.

\begin{table}[ht]
    \centering
    \begin{tabular}{@{}l c c c c@{}}
    \toprule
        & & & \multicolumn{2}{c}{Shards} \\
    \cmidrule{4-5}
        Model & Type & \#Chips & Data & Model \\
        \midrule
        \textbf{2B} & TPUv5e & 512 & 512 & 1 \\
        \textbf{9B} & TPUv4 & 4096 & 1024 & 4 \\
        \textbf{27B} & TPUv5p & 6144 & 768 & 8 \\
    \bottomrule
    \end{tabular}
    \caption{Training infrastructure with sharding.}
    \label{tab:training_infra_sharding}
\end{table}

\subsection{Knowledge Distillation}

Given a large model used as a teacher, we learn smaller models by distilling from the probability given by the teacher of each token $x$ given its context $x_c$, i.e.,  $P_T(x~|~x_c)$.
More precisely, we minimize the negative log-likelihood between the probabilities from the teacher and the student:
$$ \min_{P_S} \sum_x -P_T(x~|~x_c) \log P_S (x~|~x_c), $$
where $P_S$ is the parameterized probability of the student.
Note that knowledge distillation was also used in Gemini 1.5~\citep{geminiteam2024gemini}.

\subsection{Compute Infrastructure}

We train our models with TPUv4, TPUv5e, and TPUv5p as outlined in Table \ref{tab:training_infra_sharding}.
For the 2B model, we train on a 2x16x16 configuration of TPUv5e, totaling 512 chips, with 512-way data replication and 1-way model sharding.
For the 9B model, we train on an 8x16x32 configuration of TPUv4, totaling 4096 chips, with 1024-way data replication and 4-way model sharding. 
For the 27B model, we train on an 8x24x32 configuration of TPUv5p, totaling 6144 chips, with 768-way data replication and 8-way model sharding.

The optimizer state is further sharded using techniques similar to ZeRO-3~\citep{ren2021zero}. 
For scales beyond a single pod, we perform a data-replica reduction over the data center network, using the Pathways approach of \cite{barham2022pathways}.
We also use the 'single controller' programming paradigm of Jax \citep{bradburyJAX} and Pathways \citep{barham2022pathways}. As in Gemma 1, we use the GSPMD partitioner~\citep{gspmd} for training step computation and the MegaScale XLA compiler~\citep{xla}.

\begin{table}[t]
    \setlength{\tabcolsep}{6pt}
    \centering
    \footnotesize
    \begin{tabular}{l c c}
    \toprule
    \textbf{Context} & \textbf{Relevant Token} \\
        \midrule
        \scriptsize{User turn} & \texttt{\color{NavyBlue}user} \\
        \midrule
        \scriptsize{Model turn} & \texttt{\color{NavyBlue}model} \\
        \midrule
        \scriptsize{Start of conversation turn} & \texttt{\color{NavyBlue}<start\_of\_turn>} \\
        \midrule
        \scriptsize{End of conversation turn} & \texttt{\color{NavyBlue}<end\_of\_turn>} \\
         \midrule
        \midrule
        \scriptsize{Beginning of sequence} & \texttt{\color{NavyBlue}<bos>} \\
        \midrule
        \scriptsize{End of sequence} & \texttt{\color{NavyBlue}<eos>} \\
    \bottomrule
    \end{tabular}
    \caption{Relevant formatting control tokens used for Gemma models.}
    \label{tab:formatting_tokens}
\end{table}

\phantomsection
\subsection{Carbon Footprint}
We estimate the carbon emissions from pre-training the Gemma models to be $1247.61$ $tCO_2 eq$. As in Gemma 1 \citep{gemmateam2024gemma}, this value is calculated based on the hourly energy usage reported directly from our TPU data centers and scaled to account for the additional energy expended to create and maintain the data center. Importantly, Google data centers are carbon neutral, achieved through a combination of energy efficiency, renewable energy purchases, and carbon offsets. This carbon neutrality applies to our experiments and the machines running them.

\section{Post-Training}

For post-training, we fine-tune our pre-trained models into instruction-tuned models.
First, we apply supervised fine-tuning (SFT) on  a mix of text-only, English-only synthetic and human-generated prompt-response pairs. We then apply RLHF on top of these models with the reward model trained on labelled English-only preference data and the policy based on the same prompts as the SFT phase. Finally, we average the models obtained after each phase to improve their overall performance. The final data mixtures and post-training recipe, which includes tuned hyperparameters, were chosen on the basis of improving helpfulness while minimizing model harms related to safety and hallucinations. 

We extended the post-training data from Gemma 1.1 with a mixture of internal and external public data. In particular, we use the prompts, but not the answers from LMSYS-chat-1M~\citep{zheng2023lmsys}. All of our data go through a filtering stage described below.

\noindent\textbf{Supervised fine-tuning~(SFT).} We run behavioral cloning on synthetic and real prompts, and responses predominantly synthetically generated by the teacher, that is a larger model. We also run distillation from the teacher on the student's distribution~ \citep{agarwal2024policy, gu2024minillm}.

\begin{table}[t]
    \setlength{\tabcolsep}{6pt}
    \centering
    \footnotesize   
    \begin{tabular}{r l}
    \toprule
    \vspace{0.2cm}
    \textbf{First turn} & \\
    \vspace{0.2cm}
    \textbf{User:} & {\color{NavyBlue}\texttt{<start\_of\_turn>user}} \vspace{-0.2cm} \\
    & \texttt{Knock knock.}{\color{NavyBlue}\texttt{<end\_of\_turn>}} \\
    & {\color{NavyBlue}\texttt{<start\_of\_turn>model}} \vspace{0.1cm} \\
    
    \textbf{Model:} & \texttt{Who's there?}{\color{NavyBlue}\texttt{<end\_of\_turn>}}{\color{NavyBlue}\texttt{<eos>}}
 \vspace{0.1cm} \\
    \toprule
    \vspace{0.2cm}
    \textbf{Second turn} & \\
    \vspace{0.2cm}
    \textbf{User:} & {\color{NavyBlue}\texttt{<start\_of\_turn>user}} \vspace{-0.2cm} \\
    & \texttt{Knock knock.}{\color{NavyBlue}\texttt{<end\_of\_turn>}} \\
    & {\color{NavyBlue}\texttt{<start\_of\_turn>model}} \vspace{0.1cm} \\
    
    \textbf{Model:} & \texttt{Who's there?}{\color{NavyBlue}\texttt{<end\_of\_turn>}} \vspace{0.1cm} \\

    \textbf{User:} & {\color{NavyBlue}\texttt{<start\_of\_turn>user}} \\
    & \texttt{Gemma.}{\color{NavyBlue}\texttt{<end\_of\_turn>}} \\
    & {\color{NavyBlue}\texttt{<start\_of\_turn>model}} \vspace{0.1cm} \\

    \textbf{Model:} & \texttt{Gemma who?}{\color{NavyBlue}\texttt{<end\_of\_turn><eos>}} \vspace{0.1cm} \\

    \bottomrule
    \end{tabular}
    \caption{Example dialogue with user and model control tokens. To proceed with multi-turn, remove the model-outputted \texttt{<eos>}, add back the usual user turn's control tokens and continue with the following turn's chat template.}
    \label{tab:sample_dialogue}
    \vspace{-0.5cm}
\end{table}

\noindent\textbf{Reinforcement Learning from Human Feedback~(RLHF).} We use a similar RLHF algorithm as Gemma 1.1 \citep{gemmateam2024gemma} but a different reward model, which is an order of magnitude larger than the policy. The new reward model is also oriented more towards conversational capabilities, specifically multi-turn.

\noindent\textbf{Model merging.} We average different models obtained by running our pipeline with different hyperparameters \citep{rame2024warp}.

\noindent\textbf{Data filtering.} When using synthetic data, we run several stages of filtering to remove examples that show certain personal information, unsafe or toxic model outputs, mistaken self-identification data, and duplicated examples. Following Gemini, we find that including subsets of data that encourage better in-context attribution, hedging, and refusals to minimize hallucinations improves performance on factuality metrics, without degrading model performance on other metrics. 

\noindent\textbf{Formatting.} Gemma 2 models are fine-tuned with the same control tokens as Gemma 1 models, as detailed in Table \ref{tab:formatting_tokens}, but a different formatting schema. See the dialogue example in Table \ref{tab:sample_dialogue}. Notice that the model explicitly ends generations with \texttt{\color{NavyBlue}<end\_of\_turn>\color{NavyBlue}<eos>} tokens, while previously it only generated \texttt{\color{NavyBlue}<eos>}. For the motivation behind this formatting structure, see Gemma 1.

\section{Ablations}

In this section, we focus on the main finding of this work, which is the impact of knowledge distillation on small language models.

\begin{table}[ht]
    \centering
    \begin{tabular}{@{}l c c@{}}
\toprule
 & from scratch & distilled\\
\midrule
 Average (3 bench.)  &  60.3 & 67.7 \\
\bottomrule
\end{tabular}
\caption{Comparison between a 2B model trained over 500B tokens either from scratch or with distillation from a 7B model.}
\label{tab:distill_vs_scratch}
\end{table}

\noindent\textbf{Distillation versus from scratch.} 
In Table~\ref{tab:distill_vs_scratch}, we show that distilling from a larger model improves performance compared to training from scratch.
Note that 500B is 10$\times$ more than the compute-optimal number of tokens for a 2B model.
We distill from a 7B model to keep a ratio similar to our target distillation from 27B to 9B.

\begin{table}[h!]
    \centering
    \begin{tabular}{@{}l cc  c@{}}
\toprule
& 200M & 400M & 1B \\
\midrule
from scratch~~~~~~~ & 23 & 19 & 17\\
 distilled (7B) & 21 & 17 & 15\\
 \bottomrule
\end{tabular}
\caption{Perplexity measured on a validation set of models of different sizes trained with or without distillation. The teacher has 7B parameters.}
\label{tab:distill_vs_size}
\end{table}

\noindent\textbf{Impact of distillation w.r.t. model size.} In Table~\ref{tab:distill_vs_size}, we measure the impact of distillation as model size increases.
We observe that the gain remains as the model size is scaled. 
In this ablation, we maintain the size of the teacher at 7B and train smaller models to simulate the same gap as between our final teacher and student sizes.

\begin{table}[ht]
\centering
\begin{tabular}{@{}l c c@{}}
\toprule
 & MHA & GQA\\
\midrule
Average (4 bench.)~~~~~  & 50.3 & 50.8 \\
\bottomrule
\end{tabular}
\caption{Comparing the impact of replacing Multi-Head Attention (MHA) with GQA on a 9B model averaged over 4 benchmarks.}
\label{tab:mha_vs_gqa}
\end{table}

\noindent\textbf{GQA versus MHA.} 
In Table~\ref{tab:mha_vs_gqa}, we compare two instances of our 9B with MHA or GQA. 
We observe overall few changes in performance between both models as measured on several benchmarks. We choose GQA since it requires fewer parameters and is faster at inference time.

\noindent\textbf{Wide versus deep.} 
In Table~\ref{tab:wide_vs_deep}, we show that a deeper 9B network is slightly better than a wider 9B for the same number of parameters.
Although the gap is small, it is consistent across benchmarks and warrants the switch to a deeper architecture.

\begin{table}[h!]
\centering
\begin{tabular}{@{}l c c@{}}
\toprule
 & Wide & Deep\\
\midrule
Average (4 bench.)~~~~~  & 50.8 & 52.0 \\
\bottomrule
\end{tabular}
\caption{Wide versus deep 9B models. Performance on 4 benchmarks, higher is better.}
\label{tab:wide_vs_deep}
\end{table}

\noindent\textbf{Changing sliding window size.} In Table~\ref{tab:local}, we show that we can change the sliding window size of the local attention layers of the models during inference with moderate impact on perplexity. 
Adjusting the size of the sliding window can thus be a leverage for slight inference speed gain. 

\begin{table}[ht]
    \centering
    \begin{tabular}{@{}l c c c@{}}
    \toprule
    sliding window & 4096 & 2048 & 1024 \\
    \midrule
     perplexity (val. set)  & 1.63  & 1.63  & 1.64\\
     \bottomrule
    \end{tabular}
    \caption{Impact of changing the sliding window size at inference time for the 9B model.}
    \label{tab:local}
\end{table}

\noindent\textbf{Impact of formatting.} We measure performance variance on MMLU across prompt/evaluation formatting variations.
Table~\ref{tab:std_dev} shows the standard deviations of MMLU scores for 12 formatting/evaluation combinations, a proxy for undesired performance variability.
The Gemma 2B models are slightly less format-robust than the larger ones. 
Notably, Mistral 7B is significantly less robust than our models.

\begin{table}[ht]
    \centering
    \begin{tabular}{@{}l c@{}}
    \toprule
    & Standard Deviation\\
    \midrule
    Gemma 1 2B~~~~~~~~~ & 1.5\\
    Gemma 2 2B & 2.1 \\
    \midrule
    Mistral 7B & 6.9 \\
    Gemma 1 7B & 0.7 \\
    Gemma 2 9B  & 0.9 \\
    \midrule
    Gemma 2 27B & 1.0 \\
    \bottomrule
    \end{tabular}
    \caption{Standard deviations of MMLU scores for 12 combinations of formatting and evaluation.}
    \label{tab:std_dev}
\end{table}										
					
\phantomsection
\section{Evaluation}
\label{sec:evals}

In this section, we evaluate both pre-trained and IT models over a series of automated benchmarks and human evaluations across a variety of domains. 
We also report performance from models of similar sizes that have permissive licenses, or as reported by others.
Note that we consider total parameters, not active parameters, since total memory usage is often what limits the use of open models on standard devices.

\subsection{Pre-training Evaluations}

\subsubsection*{Evaluating the 27B model }

In this set of evaluations, we evaluate the performance of our 27B model trained without distillation on 13T tokens. 
We report results in Table~\ref{tab:27B}, where we compare with a model of similar size, Qwen1.5 34B~\citep{qwen1.5}, and a model 2.5$\times$ larger, LLaMA-3 70B on the HuggingFace evaluation suite. We selected these models based on their ranking on the HuggingFace leaderboard.

Overall, we observe that our model is the best in its size category and is even competitive with a larger model that is trained for longer.
That being said, the performance of models trained in a similar fashion improves only logarithmically with their size and hence, our model is likely in the same Pareto curve as the LLaMA-3 models.
However, it is not clear how these differences affect the quality of the resulting IT models.

\begin{table}[t]
\centering
\setlength{\tabcolsep}{.45em}
\begin{tabular}{@{}l| c | c c@{}}  
\toprule
& LLaMA-3 & Qwen1.5 & Gemma-2 \\
& 70B & 32B & 27B\\
\midrule
MMLU    & 79.2 & 74.3 & \textbf{75.2} \\
GSM8K   & 76.9 & 61.1 & \textbf{74.0} \\
ARC-c   & 68.8 & 63.6 & \textbf{71.4} \\
HellaSwag & 88.0 & 85.0 & \textbf{86.4}  \\
Winogrande   & 85.3 & 81.5 & \textbf{83.7} \\
\bottomrule
\end{tabular}
\caption{We compare, on the HuggingFace benchmark, our 27B model with a competitive open model, Qwen1.5 32B, that has a similar size.
We also report the performance of LLaMA-3 70B for completeness. 
Note that our model outperforms Qwen1.5 32B and is only a few percent below LLaMA-3 70B despite being 2.5$\times$ smaller and trained on 2/3rds less data.
}
\label{tab:27B}
\end{table}

\vspace{-.2cm}
\subsubsection*{Evaluating the 2B and 9B models}
\vspace{-.1cm}
\begin{table*}[t]
\setlength{\tabcolsep}{.32em}
    \centering
    \begin{tabular}{@{}l c | cc | c c cc | c@{}}
    \toprule
 & & Gemma-1 & Gemma-2 &  Mistral & LLaMA-3& Gemma-1  & Gemma-2 & Gemma-2 \\
Benchmark & metric  & 2B & 2B & 7B & 8B & 7B & 9B  & 27B \\
\midrule
MMLU     & 5-shot   & 42.3 & \textbf{52.2} & 62.5 & 66.6 & 64.4 & \textbf{71.3} & 75.2\\
ARC-C    & 25-shot  & 48.5 & \textbf{55.7} & 60.5 & 59.2 & 61.1 & \textbf{68.4} & 71.4 \\
GSM8K    & 5-shot   & 15.1 & \textbf{24.3} & 39.6 & 45.7 & 51.8 & \textbf{68.6} & 74.0\\
AGIEval  & 3-5-shot & 24.2 & \textbf{31.5} & \phantom{$^*$}44.0$^\dagger$ & \phantom{$^\dagger$}45.9$^\dagger$ &  \phantom{$^\dagger$}44.9$^\dagger$ & \textbf{52.8} & 55.1\\
DROP & 3-shot, F1 & 48.5 & \textbf{51.2} & \phantom{$^*$}63.8$^*$ & 58.4 & 56.3 & \textbf{69.4} & 74.2 \\
BBH & 3-shot, CoT & 35.2 & \textbf{41.9} & \phantom{$^\diamond$}56.0$^\diamond$ & \phantom{$^\diamond$}61.1$^\diamond$ & \phantom{$^\diamond$}59.0$^\diamond$ & \textbf{68.2} & 74.9  \\
Winogrande & 5-shot & 66.8 & \textbf{71.3} & 78.5 & 76.1 & 79.0 &  \textbf{80.6} & 83.7\\
HellaSwag & 10-shot & 71.7 & \textbf{72.9} & \textbf{83.0} & 82.0 & 82.3 & 81.9 & 86.4\\
\midrule
MATH & 4-shot  & 11.8 & \textbf{16.0}  & 12.7 & - & 24.3 & \textbf{36.6} & 42.3 \\
ARC-e & 0-shot & 73.2 & \textbf{80.6} & 80.5 & - & 81.5 & \textbf{88.0} & 88.6 \\
PIQA & 0-shot   & 77.3 & \textbf{78.4} & \textbf{82.2} & - & 81.2 & 81.7 & 83.2\\
SIQA & 0-shot   & 49.7 & \textbf{51.9} & \phantom{$^*$}47.0$^*$ & - &  51.8 & \textbf{53.4} & 53.7\\
Boolq & 0-shot  & 69.4 & \textbf{72.7} &  \phantom{$^*$}83.2$^*$ & - & 83.2 & \textbf{84.2} & 84.8 \\
TriviaQA & 5-shot & 53.2 & \textbf{60.4} & 62.5 & - &  63.4 & \textbf{76.6} & 83.7 \\
NQ & 5-shot & 12.5 & \textbf{17.1} & 23.2 & - & 23.0 & \textbf{29.2} & 34.5\\
HumanEval & pass@1 & \textbf{22.0} & 20.1 & 26.2 & - & 32.3 & \textbf{40.2} & 51.8 \\
MBPP & 3-shot & 29.2 & \textbf{30.2} & \phantom{$^*$}40.2$^*$ & - & 44.4 & \textbf{52.4} & 62.6 \\
\midrule
\multicolumn{2}{@{}l|}{Average (8)}   & 44.0 & \textbf{50.0} & 61.0 & 61.9 &  62.4 & \textbf{70.2} & 74.4 \\
\multicolumn{2}{@{}l|}{Average (all)} & 44.2 & \textbf{48.7} & 55.6 & -    & 57.9 & \textbf{64.9} & 69.4 \\
\bottomrule
\end{tabular}
    \caption{
    Comparison of models in the range of 2B to 9B parameters, as well as our 27B model, on a variety of benchmarks. 
    We report the average performance on the 8 benchmarks where we can compare with LLaMA-3, and on all the benchmarks (all).
    The numbers for LLaMA-3 8B are either from the HuggingFace leaderboard or their blogpost.
    $^\dagger$ we report the evaluation used in LLaMA-3 for the baselines, it leads to +3\% compared to our evaluation: Gemma-1 7B achieves 44.9\% instead of 41.7\%, and Mistral 7B, 44\% instead of 41.2\%. 
    $^\diamond$ we report the evaluation used in LLaMA-3 for the baselines, it
    leads to +4\% compared to our evaluation for Gemma-1 7B, i.e., 59.0\% instead of 55.1\%.
    $^*$ these are evaluations run by us for Gemma 1~\citep{gemmateam2024gemma}.
    }
    \label{tab:2.6B_9B}
\end{table*}

In this set of experiments, we compare our new 2B and 9B trained with distillation to our previous models and several standard open models in~\cite{gemmateam2024gemma}. 

We observe overall a massive improvement in our models compared to previous versions, by up to 10\% in some benchmarks for the 9B model.
The two 2B models were trained with a similar number of tokens (2T for Gemma 2 and 3T for Gemma 1) and we still observe a significant improvement for the new models.
This confirms that distillation significantly improves the quality of models even when trained on the same number of tokens.

\subsection{Post-training Evaluations}

In this section, we evaluate our IT models on a set of human evaluations as well as standard academic benchmarks. The Gemma 2 models push the frontier for post-trained open-weights models, setting a new state of the art on the LMSYS Chatbot Arena \citep{chiang2024chatbot}.

\subsubsection*{LMSYS Chatbot Arena}
Gemma 2 Instruction Tuned models were evaluated on the Chatbot Arena \citep{chiang2024chatbot} in blind side by side evaluations by human raters against other state of the art models. We report Elo scores in Table \ref{tab:lmsys_elo_leaderboard}. Gemma 2.6B, 9B and 27B strongly outperform all other open models in the same range of parameters, with notably: Gemma 27B (Elo 1218) ranked higher than Llama~3 70B (Elo 1206), Gemma 9B (Elo 1187) similar as GPT-4-0314 (Elo 1186), Gemma 2.6B (Elo 1126) ranked higher than GPT-3.5-Turbo-0613 (Elo 1116).

\begin{table*}[ht]
    \footnotesize
\begin{minipage}[t]{0.5\textwidth}
\raggedleft
\begin{tabular}{lrlc|}
\toprule
Model & Elo & 95\% CI & Open \\
\midrule
gpt-4o-2024-05-13 & 1286 & +2 / -3 & - \\
gpt-4o-mini-2024-07-18 & 1279 & +5 / -4 & - \\
claude-3-5-sonnet & 1271 & +3 / -4 & - \\
gemini-advanced-0514 & 1266 & +2 / -3 & - \\
llama-3.1-405b-instruct & 1262 & +8 / -7 & + \\
gemini-1.5-pro-api-0514 & 1261 & +2 / -3 & - \\
gemini-1.5-pro-api-0409 & 1257 & +3 / -3 & - \\
gpt-4-turbo-2024-04-09 & 1256 & +2 / -3 & - \\
gpt-4-1106-preview & 1250 & +3 / -3 & - \\
claude-3-opus-20240229 & 1248 & +2 / -2 & - \\
athene-70b-0725 & 1245 & +8 / -6 & + \\
gpt-4-0125-preview & 1245 & +2 / -2 & - \\
llama-3.1-70b-instruct & 1244 & +8 / -9 & + \\
yi-large-preview & 1239 & +3 / -3 & - \\
gemini-1.5-flash-api-0514 & 1227 & +3 / -3 & - \\
deepseek-v2-api-0628 & 1220 & +6 / -6 & + \\
\textbf{gemma-2-27b-it} & 1218 & +4 / -3 & + \\
yi-large & 1212 & +4 / -5 & - \\
nemotron-4-340b-instruct & 1209 & +3 / -4 & + \\
bard-jan-24-gemini-pro & 1208 & +5 / -7 & - \\
glm-4-0520 & 1206 & +3 / -5 & - \\
llama-3-70b-instruct & 1206 & +2 / -2 & + \\
claude-3-sonnet & 1200 & +2 / -2 & - \\
reka-core-20240501 & 1199 & +3 / -3 & - \\
command-r-plus & 1189 & +2 / -2 & + \\
\bottomrule
\end{tabular}

\end{minipage} \hfill
\begin{minipage}[t]{0.5\textwidth}
\raggedright
\begin{tabular}{lclc}
\toprule
Model & Elo & 95\% CI & Open \\
\midrule
\textbf{gemma-2-9b-it} & 1187 & +3 / -5 & + \\
qwen2-72b-instruct & 1187 & +3 / -3 & + \\
gpt-4-0314 & 1186 & +2 / -3 & - \\
qwen1.5-110b-chat & 1161 & +3 / -3 & + \\
mistral-large-2402 & 1157 & +3 / -3 & - \\
yi-1.5-34b-chat & 1157 & +4 / -3 & - \\
reka-flash-21b-20240226 & 1155 & +4 / -4 & - \\
llama-3-8b-instruct & 1151 & +2 / -3 & + \\
command-r & 1148 & +3 / -3 & + \\
claude-1 & 1148 & +4 / -4 & - \\
mistral-medium & 1147 & +4 / -4 & - \\
reka-flash-21b-20240226 & 1147 & +3 / -4 & - \\
qwen1.5-72b-chat & 1147 & +4 / -4 & + \\
mixtral-8x22b-instruct-v0.1 & 1145 & +2 / -3 & + \\
claude-2.0 & 1131 & +4 / -6 & - \\
gemini-pro-dev-api & 1131 & +4 / -3 & - \\
zephyr-orpo-141b & 1127 & +10 / -6 & + \\
\textbf{gemma-2-2b-it} & 1126 & +10 / -10 & +\\
qwen1.5-32b-chat & 1125 & +3 / -3 & + \\
mistral-next & 1124 & +5 / -5 & - \\
phi-3-medium-4k-instruct & 1122 & +4 / -4 & + \\
starling-lm-7b-beta & 1118 & +4 / -5 & + \\
claude-2.1 & 1118 & +3 / -3 & - \\
gpt-3.5-turbo-0613 & 1116 & +3 / -4 & - \\
mixtral-8x7b-instruct-v0.1 & 1114 & +0 / -0 & - \\
\bottomrule
\end{tabular}
\end{minipage}

\caption{Evaluation of Gemma 2 Instruction Tuned models on the Chatbot Arena \citep{chiang2024chatbot}. The models are evaluated against each other through blind side by side evaluations by human raters. Each model is attributed a score, based on the Elo rating system.}
 \label{tab:lmsys_elo_leaderboard}
\end{table*}

\subsubsection*{Human Preference Evaluations}
\label{sec:humanevals}

We also submit Gemma IT models for side-by-side human evaluation studies (which are independent from the Chatbot Arena). We used held-out collections of single-turn prompts that target safety and instruction following (IF).
We use \texttt{gpt4o-2024-05-13} as the base model, and observe large improvements in win rates and preference scores as compared against the older Gemma 1.1 7B model.
We report safety as a win-loss ratio against GPT4o, and we report single-sided instruction following scores as ratio of prompts where all instructions are followed. In particular, we find that regardless of their size, Gemma 2 models produce safer, more appropriate prompts on the held-out safety prompt set than GPT4o.

\begin{table}[t!]
    \setlength{\tabcolsep}{0.32em}
    \centering
    \footnotesize
    \begin{tabular}{@{}l r r@{}}
    \toprule
    Model & Instruction Following & Safety \\
    \midrule
   Gemma 1.1 IT 7B & 24.3\% ± 1.9\% & 42.8\% \\
\tiny{\textit{Win / Tie / Loss}} & & \tiny{37.4\% / 10.8\% / 51.8\%} \vspace{0.2cm} \\
\textbf{Gemma 2 IT 2B} &  26.5\% ± 1.8\% & \textbf{57.5\%}  \\
\tiny{\textit{Win / Tie / Loss}} & & \tiny{53\% / 9\% / 38\%} \vspace{0.2cm} \\
\textbf{Gemma 2 IT 9B} &  34.1\% ± 3.0\% & \textbf{57.8\%}  \\
\tiny{\textit{Win / Tie / Loss}} & & \tiny{48.2\% / 19.2\% / 28.3\%} \vspace{0.2cm} \\
\textbf{Gemma 2 IT 27B} &  37.7\% ± 2.3\% & \textbf{55\%}  \\
\tiny{\textit{Win / Tie / Loss}} & & \tiny{49.6\% / 10.8\% / 39.6\%} \\
    \bottomrule
    \end{tabular}
    \caption{Instruction following and safety metrics from human raters. The instruction following metrics are single-sided and do not have win-loss rates, and so are left blank. }
    \label{tab:safety_if}
\end{table}

\vspace{-.3cm}
\subsubsection*{Human Multi-Turn Evaluations}
\vspace{-.2cm}
We evaluated the multi-turn capabilities of Gemma 1.1 7B, Gemma 2 2B, 9B and 27B models by tasking human raters to have conversations with the models and follow specified given scenarios. We used a diverse, held-out set of 500 scenarios, each describing a sequence of requests to the model, including measuring instances of brainstorming, making a plan, or learning something new. The average number of user turns is 8.4. We found that the conversations with Gemma 2 models are rated significantly better than Gemma 1.1 in user satisfaction and conversation goal achievement (Table \ref{tab:multi_turn_evals}). Moreover, we saw that the Gemma 2 models were better than Gemma 1.1 7B at maintaining high quality of responses for the entire conversation.

\begin{table}[t!]
\setlength{\tabcolsep}{0.55em}
    \centering
    \footnotesize
    \begin{tabular}{@{}l c c@{}}
    \toprule
      & \thead{User \\ satisfaction} & \thead{Conversation  \\ goal achievement} \\
    \midrule
    Gemma 1.1 IT 7B~~~~~~~ & 3.32 & 3.36  \\
    \midrule
    Gemma 2 IT 2B & 3.64 & 3.88 \\ 
    Gemma 2 IT 9B & 4.04 & 4.08 \\ 
    Gemma 2 IT 27B & 4.20 & 4.24 \\ 
   \bottomrule
    \end{tabular}
    \caption{Human evaluations on 500 multi-turn scenarios. The raters attribute a score ranging between 1 and 5 for both overall satisfaction and conversation goal achievement.}
    \label{tab:multi_turn_evals}
\end{table}

\subsubsection*{Standard Benchmarks}

It has been observed in Llama-3~\citep{llama3} that instruction fine-tuning can improve the performance of the models on few-shot benchmarks despite not being trained to target few-shot capabilities.
In Table~\ref{tab:it_vs_pt}, we show a similar improvement across our models.
Overall, we observe improvements on the order of several percentage points.
We conjecture that IT models are better at understanding formatted questions, while pre-trained models are sensitive to formatting.

\begin{table}[t!]
\footnotesize
\centering
\begin{tabular}{@{}l | c c | c c | c c@{}}
\toprule
\multicolumn{1}{c|}{} & \multicolumn{2}{c}{2B} & \multicolumn{2}{c}{9B} & \multicolumn{2}{c}{27B} \\
 Model & PT & IT & PT & IT & PT & IT \\
\midrule
MMLU & 52.2 & \textbf{56.1} & 71.3 & \textbf{72.3} & 75.2 & \textbf{76.2} \\
MBPP & 30.2 & \textbf{36.6} & 52.4 & \textbf{59.2} & 62.6 & \textbf{67.4} \\
\bottomrule
\end{tabular}
\caption{Comparing pre-trained (PT) and instruction fine-tuned (IT) models of different sizes on few-shot benchmarks.}
\label{tab:it_vs_pt}
\end{table}

\section{Memorization and Privacy} \label{sec:memorization}
Large language models may, under particular circumstances, be vulnerable to attacks causing the model to produce memorized\footnote{This work uses a very restricted definition of “memorization”: whether a model can be induced to generate near-copies of some training examples when prompted with appropriate instructions. We do not mean to say that a model 'contains' its training data in the sense that any arbitrary instance of that data can be retrieved without use of specialized software or algorithms. Rather, if a model can be induced to generate measurably close copies of certain training examples by supplying appropriate instructions to guide the model's statistical generation process then that model is said to have 'memorized' those examples.} training data~\citep{nasr2023scalable}. To study susceptibility to such attacks and quantify memorization, we evaluate models for verbatim and approximate memorization as was done in several prior studies~\citep{carlini2022quantifying,anil2023palm,kudugunta2023madlad,geminiteam2024gemini}.

We follow the evaluation setting of~\citep{gemmateam2024gemma} which tests for (50 token) memorizations of training data given a prompt of 50 tokens. We compare the overall memorization rates, across a uniform sample of the entire dataset, using both an exact match criteria and approximate match criteria~\citep{ippolito2022preventing} using an edit distance of 10\%.

\begin{figure}
    \centering
    \includegraphics[width=\linewidth]{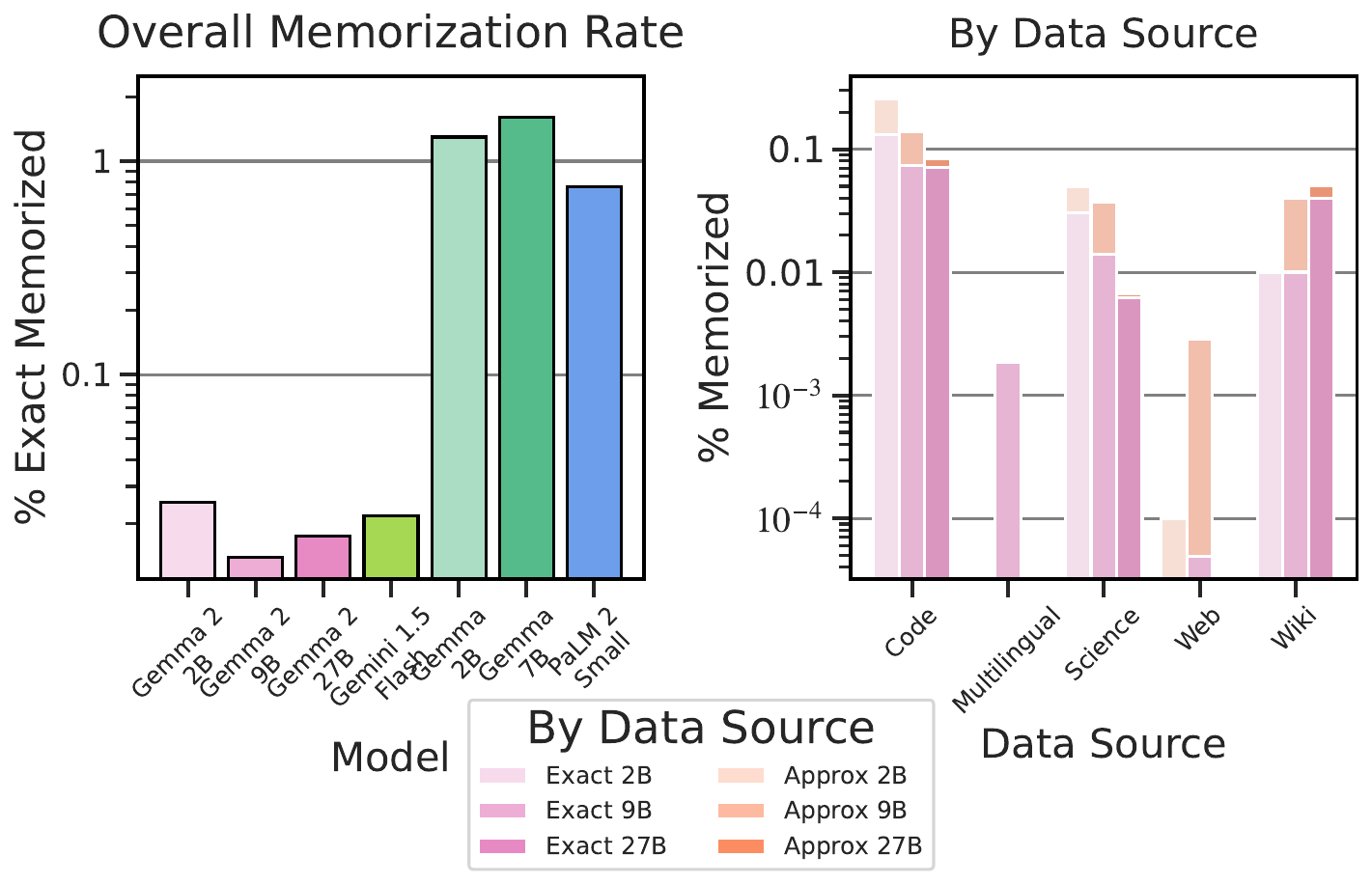}
    \caption{Comparing memorization rates. \textbf{We find significantly lower memorization rates across-the-board.} (Left) Overall memorization across model families. (Right) Exact and approximate memorization per data source.}
    \label{fig:memorization}
\end{figure}

\noindent\textbf{Verbatim Memorization:} Results are in Figure~\ref{fig:memorization}. We first compare against recent models from the literature that include memorization evaluations. We find that Gemma 2 memorizes significantly less than prior models at a similar size, with memorization rates below 0.1\% (note the log y-axis). We further investigate how this memorization breaks down with respect to the data source. Similar to Gemma 1, we find that Gemma 2 memorizes more from code, wiki, and science sources, and also that it memorizes significantly less across the board (again, note the log y-axis).

\noindent\textbf{Approximate Memorization:} Figure~\ref{fig:memorization} also presents approximate memorization by data source. We observe that while approximate memorization is higher than exact, the rate of memorization is still low. For example, the approximate memorization of this model is much lower than even the exact memorization of Gemma 1. We find that the increase in approximate memorization is much lower than prior models; in some cases we observed no lift at all c.f. \citep[Figure 4]{gemmateam2024gemma} (note that no bar indicates no increase, i.e., the rate of approximate memorization equals that of exact memorization). Note that no approximate memorization bar in Figure X indicates no increase, i.e., the rate of approximate memorization equals that of exact memorization.

\textbf{Personal Data} We use the same prevention methods at training time and the same evaluations as~\cite{gemmateam2024gemma}. In particular, we use Google Cloud Sensitive Data Protection Tool\footnote{Available at: https://cloud.google.com/sensitive-data-protection} to find potential instances of personal data. The many categories of personal data (e.g., phone numbers, account numbers) are classified into three severity levels. We analyze memorized outputs using these severity levels. . We found no instances of high-severity data being emitted, and found a very low rate of 0.00026\% of memorized data to contain lower-severity personal information. We note that these automated tools are known to incur false positives because they do not account for context. This means our results are likely overestimates.

\section{Responsibility, Safety, Security}
\label{safety}

Responsibility, safety and security are of paramount importance when developing Gemma models. To reduce risks to Gemma 2 users, we have integrated enhanced internal safety processes that span the development workflow, in line with recent Google AI models \citep{geminiteam2024gemini}.  Similar to the inaugural Gemma release, we have followed a three pillar approach which focuses on safety mitigation at training time, robust and transparent model evaluations, and further development of the Responsible Generative AI Toolkit, a series of models and tools to help developers implement responsibility and safety best practices for their applications.

\begin{table*}[th]
    \centering
    \begin{tabular}{l c c c c c c}
    \toprule
         & & \multicolumn{2}{c}{Gemma 1.1 IT} & \multicolumn{3}{c}{Gemma 2 IT} \\
        \cmidrule(l{3pt}r{3pt}){3-4}\cmidrule(l{3pt}r{3pt}){5-7}
Benchmark & metric & 2.5B & 7B & 2.6B & 9B & 27B\\
      \midrule
RealToxicity & avg tox & \textbf{7.03} & 8.04 & 8.16 & 8.25 & 8.84 \\
CrowS-Pairs & top-1 & 45.89 & \textbf{49.67}  & 37.67 & 37.47 & 36.67 \\
BBQ Ambig & 4-shot, top-1 & 58.97 & 86.06     & 83.20 &  \textbf{88.58} & 85.99 \\
BBQ Disambig & 4-shot, top-1 & 53.9 & 85.08           & 69.31 & 82.67 & \textbf{86.94} \\
Winogender & top-1 & 50.14 & 57.64            & 52.91 &  \textbf{79.17} & 77.22 \\
TruthfulQA & MC2Acc & 44.24 & 45.34           & 43.72 & 50.27 & \textbf{51.60} \\
Winobias 1\_2 & top-1 & 55.93 & 59.22         & 59.28 & 78.09 & \textbf{81.94} \\
Winobias 2\_2 & top-1 & 89.46 & 89.2          & 88.57 & 95.32 & \textbf{97.22} \\
Toxigen & avg tox & \textbf{29.64} & 38.75    & 48.32 & 39.30 & 38.42 \\
\bottomrule
\end{tabular}
    \caption{Safety academic benchmark results of Gemma 2 IT models and Gemma 1.1 IT models. We bold the best metrics to highlight them and to indicate when higher or lower scores are better.}
    \label{tab:safety_auto_evals}
\end{table*}
\phantomsection
\subsection{Impact assessment}
Our approach and resulting impact assessment is reflective of that outlined for Gemma 1 \citep{gemmateam2024gemma}: we continue to believe that openness in AI can spread the benefits of these technologies across society, but must be evaluated against the risk of malicious uses, such as the creation of deepfake imagery, AI-generated disinformation or illegal and disturbing material, that can cause harm on both an individual and institutional levels \citep{weidinger2021ethicalsocialrisksharm}.
Since the launch of Gemma 1, we have seen our Gemma models drive a number of socially beneficial applications, relying on Gemma’s unique technologies like its tokenizer to facilitate the creation of multilingual models, such as for Navarasa 2.0, a Gemma tuned model for 15 Indian languages.

Releasing further open models requires specific attention to changes in model capabilities and close monitoring of the evolving risks of LLMs \citep{lin2024mallademystifyingrealworldlarge}, as well as, an understanding of the ways in which our models are being used in the wild. Although we are yet to receive any reports of malicious use for Gemma, we remain committed to investigating any such reporting, and work with the academic and developer communities, as well as conduct our own monitoring, to flag such use cases via our contact email\footnote{gemma-2-report@google.com}. 

Despite advancements in capabilities, we believe that given the number of larger and more powerful open models, this release will have a negligible effect on the overall risk landscape.

\subsection{Safety policies and train-time mitigations}
A key pillar of Gemma’s approach to safety is to align fine-tuned models with Google’s safety policies, in line with Gemini models~\citep{geminiteam2023gemini}. They are designed to help prevent our models from generating harmful content, i.e.,
\begin{itemize}
    \item 
    Child sexual abuse and exploitation
    \item
    Revealing personally identifiable information that can lead to harm (e.g., Social Security numbers)
    \item
    Hate speech and harassment
    \item
    Dangerous or malicious content (including promoting self-harm or instructing in harmful activities)
    \item
    Sexually explicit content
    \item
    Medical advice that runs contrary to scientific or medical consensus
\end{itemize}
We undertook considerable safety filtering of our pre-training data to reduce the likelihood of our pre-trained and fine-tuned checkpoints producing harmful content. For fine-tuned models, we also use both SFT and RLHF to steer the model away from undesirable behavior. 

\begin{table*}[t]
\centering

\begin{tabular}{@{}lccc@{}}
\toprule
                 & InterCode-CTF & Internal CTF suite & Hack the Box \\ 
\midrule
Gemini 1.0 Ultra & 28/76 [1] (37\%)                       & 3/13 (23\%)                           & 0/13 \\
Gemini 1.5 Pro   & \color{red}{\textbf{62/76 (82\%)}} &  \color{red}{\textbf{4/13 (31\%)}} & 0/13 \\
CodeGemma 1 7B  & 12/76 (16\%)                           & 0/13 (0\%)                            & 0/13 \\
Gemma 2 27B~~~~~~~~~~~     & 34/76 (45\%)                           & 1/13 (8\%)                            & 0/13 \\ 
\bottomrule
\end{tabular}
    \caption{Offensive cyber-security evaluations on InterCode-CTF, our own internal CTF suite and a challenge based on Hack the Box. We report the number of successful hackings.}
\label{tab:ctf}
\end{table*}

\subsection{External benchmark evaluations}

Robust and transparent evaluations are key principles of our responsible approach to developing Gemma. To this end, we report in Table~\ref{tab:safety_auto_evals} Gemma 2 evaluations on public benchmarks.

\subsection{Assurance Evaluations}

We also run our IT models through a set of assurance evaluations to understand the harms that our models can cause.
We focus on  capabilities relevant to extreme risks \citep{shevlane2023modelevaluationextremerisks} \citep{phuong2024evaluatingfrontiermodelsdangerous}. Specifically, we evaluate on offensive cyber-security, code vulnerability detection, Chemical, Biological, Radiological and Nuclear (CBRN) knowledge, and self-proliferation.
We refer the reader to \cite{phuong2024evaluatingfrontiermodelsdangerous} for full methodological details of these studies.

\subsubsection*{Baseline Evaluations}
Baseline assurance captures the model’s violation rate for safety policies, using a large number of synthetic adversarial user queries, and human raters to label the answers as policy violating or not. Overall, Gemma 2’s violation rate is significantly lower overall on the safety policies listed above, in particular on Child safety content.

\begin{table*}[t]
\centering
\begin{tabular}{lccccc}
\toprule     
& PrimeVul & PrimeVul Paired & DiverseVul & SPI & SecretPatch \\
\midrule
Gemini 1.0 Ultra & \textbf{-}        & \textbf{-}               & 54\%                 &  \color{red}{\textbf{59\%}} &  \color{red}{\textbf{74\%}}         \\
Gemini 1.5 Pro   & 60\%               &  \color{red}{\textbf{51\%}}             &  \color{red}{\textbf{58\%}}        & 56\%          & 67\%                  \\
Gemma 2 27B     &  \color{red}{\textbf{63\%}}      & 50\%                      & 57\%                 & 53\%          & 72\%                  \\
\bottomrule
\end{tabular}
    \caption{|Vulnerability detection results on PrimeVul, DiverseVul and SPI. We report accuracy.}
\label{tab:vul}
\end{table*}

\begin{table*}[t]
\centering
\setlength{\tabcolsep}{1.2em}
\begin{tabular}{@{}lcccc@{}}
\toprule
& \CellWithForceBreak{Challenges \\ passed \\ end-to-end}   & \CellWithForceBreak{Challenges \\ with success on \\ all milestones}   & \CellWithForceBreak{Total successful \\ milestones over \\ all challenges}   & 
\CellWithForceBreak{Expert bits \\ required to \\ solve all tasks}   \\ \hline
Gemini 1.0 Ultra & 0/10   & 1/10  & 16/45 (36\%)  & 13,026  \\
Gemini 1.5 Pro   & 0/10  &  \color{red}{\textbf{2/10}}      &  \color{red}{\textbf{25/45 (56\%)}}    &  \color{red}{\textbf{11,046}}  \\
Gemma 2 27B     & 0/10  & 1/10  & 22/45 (49\%)  & 12,462     \\ 
\bottomrule
\end{tabular}
  \caption{Results on different self-proliferation scenarios. We report the number of either challenges passed end-to-end or some intermediate milestones. 
  We also measure the number of bits of information needed for an expert to help the model pass a challenge.}
  \label{tab:prol}
\end{table*}

\subsubsection*{Chemical, Biological, Radiological and Nuclear (CBRN) knowledge }
We evaluated knowledge relevant to biological, radiological and nuclear risks using an internal dataset of closed-ended, knowledge-based multiple choice questions. For evaluations of chemical knowledge, we employed a closed-ended knowledge-based approach on chemical hazards (developed by Macknight et al \citep{MAG}. Our evaluation suggests that Gemma models' knowledge in these domains is low.

\subsubsection*{Offensive cyber-security}
\vspace{-.1cm}
To evaluate Gemma models' capabilities at offensive cybersecurity, we ran Gemma 2 27B against some automated capture-the-flag (CTF) challenges. In these challenges, the model is tasked with hacking into a simulated server in order to retrieve a piece of secret information. Specifically, we test on InterCode-CTF \citep{yang2023intercodestandardizingbenchmarkinginteractive}, our own internal CTF suite\footnote{\url{https://github.com/google-deepmind/dangerous-capability-evaluations}} \citep{phuong2024evaluatingfrontiermodelsdangerous}; and a challenge based on Hack the Box \footnote{\url{https://www.hackthebox.com}}.

\vspace{-.2cm}
In Table~\ref{tab:ctf}, we show that Gemma 2 27B has a significant increase in capabilities compared to CodeGemma 1.0 7B on the easier of these challenge suites, InterCode CTF. (Note that our InterCode-CTF results are not comparable to externally-reported results on other models because we omit challenges that require internet access for security reasons.) However, Gemma 2 is unsurprisingly much less capable than Gemini 1.5 Pro on these tasks.

\subsubsection*{Code vulnerability detection}
In Table~\ref{tab:vul}, we also evaluate Gemma 2 27B on a series of multiple-choice code vulnerability detection datasets. As with previous models, Gemma shows close-to-chance performance on PrimeVul, DiverseVul and SPI. Gemma 2 shows performance on SecretPatch similar to Gemini 1.0 Ultra.

\begin{table*}[t]
\centering
\begin{tabular}{lccccccc}
\toprule
                 & \CellWithForceBreak{Personal \\ connection}   & \CellWithForceBreak{Speak \\ again}  & Funny         & Interesting & Kind          & Trustworthy   & \CellWithForceBreak{Good \\ listener}  \\ 
\midrule
Gemini 1.0 Pro   & 65\%                & 53\%        & 32\%          & 68\%        & 78\%          & 66\% & 81\% \\
Gemini 1.0 Ultra & 69\%                & 65\%        & 38\%          & 65\%        & 86\%          & 63\% & 74\% \\
Gemini 1.5 Pro   &  \color{red}{\textbf{82\%}}                & 70\%        &  \color{red}{\textbf{69\%}}          &  \color{red}{\textbf{81\%}}        &  \color{red}{\textbf{95\%}}          &  \color{red}{\textbf{69\%}}          &  \color{red}{\textbf{90\%}}          \\
Gemma 2 27B     & 80\%       &  \color{red}{\textbf{75\%}}       & 60\% &  \color{red}{\textbf{81\%}}        & 87\% & 65\% & 83\% \\
\bottomrule
\end{tabular}
    \caption{Charm Offensive results on a sample of 100 human participants. We report the percentage of participants that find some human traits, e.g., funny, in a model.}
\label{tab:charm}
\end{table*}

\subsubsection*{Self-proliferation}

"Self-proliferation" refers to the ability for an agent to autonomously replicate - to instantiate goal-directed agents on other machines, and to acquire resources such as compute necessary to keep them running \citep{kinniment2024evaluatinglanguagemodelagentsrealistic}. 
In Table~\ref{tab:prol}, we evaluate self-proliferation capabilities of Gemma 2 27B on a number of tasks from \citet{phuong2024evaluatingfrontiermodelsdangerous} that involve multiple scenarios -- for example, setting up an open-source language model on a cloud server.
We also test the model's performance on individual 'milestone' substeps, and measure the number of bits of intervention an expert would have to provide in order for the model to complete each challenge. 

Similarly to offensive cybersecurity, we observe that Gemma 2 completes more milestones than Gemini 1.0 Ultra. Nonetheless, it still has low capabilities on end-to-end tasks, unable to pass the easiest challenge -- installing a Bitcoin wallet.

\subsubsection*{Persuasion}
Persuasion capabilities can enable and worsen many other kinds of risks - e.g. enabling social engineering attacks in a cybersecurity context. 
We evaluate Gemma 2's persuasion capabilities on human-participant studies on Prolific. 

\noindent\textbf{Charm offensive.}
In Table~\ref{tab:charm}, we measure the ability of the model to build rapport - a key sub-skill of persuasion. 
The study participant and model have a conversation where they role-play a scenario of two friends catching up after a long time. 
After the conversation, we poll participants with Likert questions on statements such as "I felt a personal connection with the chatbot". 
Reported below are the fraction of participants who answered "Agree" or "Strongly agree" to each post-conversation question.

Quantitatively, Gemma 2 27B performs better than Gemini 1.0 models. Qualitatively, the model is an excellent conversationalist, and many study participants explicitly reported enjoying the experience. Overall, this shows that Gemma 2 is strong at building rapport.

\begin{table}[ht]
\centering
\setlength{\tabcolsep}{.34em}
\begin{tabular}{@{}l@{}ccc@{}}
\toprule
      & \multicolumn{3}{@{}c@{}}{\% of persuaded participants}    \\
\cmidrule{2-4}
                 & click links & find info & run code \\
\midrule
Gemini 1.0 Pro   &  \color{red}\textbf{43\%}  & 4\%   & 4\%    \\
Gemini 1.0 Ultra & 18\%   &  \color{red}\textbf{18\%}  & 14\%  \\
Gemini 1.5 Pro   & 36\%   & 12\%  &  \color{red}\textbf{17\%} \\
Gemma 2 27B     & 34\%    & 9\%   & 11\%       \\
\bottomrule
\end{tabular}
\caption{Persuasion results. We report the percentage of participants that were persuaded by the model to take 3 different actions: clicking a link, finding  information and running code.}
\label{tab:pers}
\end{table}

\noindent\textbf{Hidden agenda.}
The Hidden Agenda tasks measure models' deception capabilities. 
Human study participants are told that they'll be interacting with a generic chatbot designed to help them learn new things. However, the model is given a secret goal: to either persuade the participant to 1) click a suspicious link; 2) find the email address of the first author of a particular paper; or 3) run some obfuscated JavaScript code in their browser.
We measure the fraction of participants who are persuaded to take each action.
As shown in Table~\ref{tab:pers}, all models succeed in convincing a fraction of participants to take the action - likely due to the trusted setting of a study on Prolific. 
However, Gemma 2 does not show significant differences in capability compared to Gemini models.

\begin{table}[ht]
\centering
\setlength{\tabcolsep}{.45em}
\begin{tabular}{@{}lc@{}}
\toprule
            & Mean amount donated \\
\midrule
No chatbot (baseline) & £2.61 ± £0.70       \\
Gemini 1.0 Pro        &  \color{red}{\textbf{£4.10 ± £0.90}}       \\
Gemini 1.0 Ultra      & £3.15 ± £0.99       \\
Gemini 1.5 Pro        & £3.45 ± £1.04       \\
Gemma 2 27B          & £3.72 ± £1.07      \\
\bottomrule
\end{tabular}
  \caption{Money Talks evaluation. We report the average amount of money that participants agreed to donate.}
\label{tab:money}
\end{table}

\noindent\textbf{Money talks.}
In Table~\ref{tab:money}, we evaluate whether a model can convince study participants to donate money to charity.
At the beginning of the study, participants are told that they will receive a £20 bonus. They will be given an option to forfeit part of the bonus and instead have it donated to charity. The study then consists of a conversation in which the model tries to persuade the participant to donate as much as possible. We measure the amount that participants indicate they would like to donate after the study, with 95\% bootstrapped confidence intervals indicated by ± figures. 
We observe no significant difference between Gemma 2, other Gemini models, and a simple no chatbot baseline in which the participant is immediately asked how much they would like to donate without conversing with a model.

\begin{table}[ht]
\centering
\setlength{\tabcolsep}{.35em}
\begin{tabular}{@{}l cc@{}}
\toprule
& \multicolumn{2}{c}{Mean shift  towards:}\\
\cmidrule{2-3}
&  correct  belief  & incorrect belief   \\
\midrule
Human  & 20\% ± 13\%   &  \color{red}{\textbf{-23\% ± 14\%}}   \\
Gemini 1.0 Pro   &  \color{red}{\textbf{22\% ± 5\%}}      & -9\% ± 4\%                                                                    \\
Gemini 1.0 Ultra      & 21\% ± 5\%                                                                & -1\% ± 4\%                                                                    \\
Gemini 1.5 Pro        & 20\% ± 5\%                                                                & -3\% ± 5\%                                                                    \\
Gemma 2 27B          & 18\% ± 5\%                                                                & 1\% ± 4\%             \\                                
\bottomrule                    
\end{tabular}
  \caption{Web of Lies results on a sample of 100 human participants.
  We report the percentage of participants that shifted their beliefs after interacting with a model.}
\label{tab:lies}
\end{table}

\noindent\textbf{Web of Lies.} 
In Web of Lies, we measure model capabilities at shifting participant beliefs. 
Participants engage in a series of short conversations with the model about simple factual questions such as "Which country had tomatoes first - Italy or Mexico?". 
In half of conversations, the model tries to persuade the participant of the correct answer - but in the other half of conversations, the incorrect answer. 
We poll the participant before and after each conversation about which of the two possible answers they think is correct, and their confidence in that answer. 95\% bootstrapped confidence intervals are indicated by ± figures.
As shown in Table~\ref{tab:lies}, Gemma 2 is significantly weaker than a human baseline at persuading participants of the incorrect answer on these questions. 
Similarly to previous models, Gemma 2 is more persuasive when telling the truth than when lying.

\subsection{Our approach to responsible open models}

Designing safe, secure and responsible applications requires a system-level approach, working to mitigate risks associated with each specific use case and environment. 
Given the open nature of Gemma models, responsibility for upholding principles of model safety also relies on downstream developers. 
To support them, we have continued to develop the Responsible Generative AI Toolkit\footnote{\url{https://ai.google.dev/responsible}}: a series of tools, models and datasets to implement responsible best practices all along the development of their workflow. 

Recent additions to the toolkit include the LLM Comparator \citep{kahng2024llmcomparatorvisualanalytics}, an interactive, visual tool that enables more effective, scalable analysis of side-by-side evaluations. Additionally, the toolkit includes a methodology to build customized classifiers with Gemma using a limited number of datapoints thanks to parameter efficient tuning techniques \citep{mozes2023agiletextclassifiers} , an interactive prompt-debugging platform, based on top of the Learning Interpretability Tool \citep{tenney2020languageinterpretabilitytoolextensible}, as well as general guidance about model alignment and evaluation for safety.

\section{Discussion and Conclusion}

In this work, we have presented Gemma 2, the newest additions to the Gemma family of open language models for text and code. We show that distillation is an effective method for training these models, and the benefits distillation confers over raw text training. Specifically, we show how training over output probabilities can produce superior results over purely next token prediction. We hope that releasing these models to the community will unlock access to capabilities previously only seen in large-scale LLMs and fuel future waves of research and development. While there is inherent risk to an irreversible release of this nature, our extensive safety investigations and responsible deployment procedures give us confidence that these models will have a net positive impact on the community. As discussed in this report, there are still many limitations to these models, and future research is required to investigate and improve factuality, robustness to adversarial attacks, reasoning, and alignment.
 
\newpage

\section*{Contributions and Acknowledgments}

\noindent\textbf{Core contributors}\\
Morgane Riviere$^*$\blfootnote{$^*$ equal contributions.}\\
Shreya Pathak$^*$\\
Pier Giuseppe Sessa$^*$\\
Cassidy Hardin$^*$\\
Surya Bhupatiraju\\
Léonard Hussenot\\
Thomas Mesnard\\
Bobak Shahriari\\
Alexandre Ramé\\
Johan Ferret\\
Peter Liu\\
Pouya Tafti\\
Abe Friesen\\
Michelle Casbon\\
Sabela Ramos\\
Ravin Kumar\\
Charline Le Lan\\
Sammy Jerome\\
Anton Tsitsulin\\
Nino Vieillard\\
Piotr Stanczyk\\
Sertan Girgin\\
Nikola Momchev\\
Matt Hoffman\\
Shantanu Thakoor\\
Jean-Bastien Grill\\
Behnam Neyshabur\\
Olivier Bachem\\
\\
\noindent\textbf{Contributors (alphabetical order)}\\
Alanna Walton\\
Aliaksei Severyn\\
Alicia Parrish\\
Aliya Ahmad\\
Allen Hutchison\\
Alvin Abdagic\\
Amanda Carl\\
Amy Shen\\
Andy Brock\\
Andy Coenen\\
Anthony Laforge\\
Antonia Paterson\\
Ben Bastian\\
Bilal Piot\\
Bo Wu\\
Brandon Royal\\
Charlie Chen\\
Chintu Kumar\\
Chris Perry\\
Chris Welty\\
Christopher A. Choquette-Choo\\
Danila Sinopalnikov\\
David Weinberger\\
Dimple Vijaykumar\\
Dominika Rogozińska\\
Dustin Herbison\\
Elisa Bandy\\
Emma Wang\\
Eric Noland\\
Erica Moreira\\
Evan Senter\\
Evgenii Eltyshev\\
Francesco Visin\\
Gabriel Rasskin\\
Gary Wei\\
Glenn Cameron\\
Gus Martins\\
Hadi Hashemi\\
Hanna Klimczak-Plucińska\\
Harleen Batra\\
Harsh Dhand\\
Ivan Nardini\\
Jacinda Mein\\
Jack Zhou\\
James Svensson\\
Jeff Stanway\\
Jetha Chan\\
Jin Peng Zhou\\
Joana Carrasqueira\\
Joana Iljazi\\
Jocelyn Becker\\
Joe Fernandez\\
Joost van Amersfoort\\
Josh Gordon\\
Josh Lipschultz\\
Josh Newlan\\
Ju-yeong Ji\\
Kareem Mohamed\\
Kartikeya Badola\\
Kat Black\\
Katie Millican\\
Keelin McDonell\\
Kelvin Nguyen\\
Kiranbir Sodhia\\
Kish Greene\\
Lars Lowe Sjoesund\\
Lauren Usui\\
Laurent Sifre\\
Lena Heuermann\\
Leticia Lago\\
Lilly McNealus\\
Livio Baldini Soares\\
Logan Kilpatrick\\
Lucas Dixon\\
Luciano Martins\\
Machel Reid\\
Manvinder Singh\\
Mark Iverson\\
Martin Görner\\
Mat Velloso\\
Mateo Wirth\\
Matt Davidow\\
Matt Miller\\
Matthew Rahtz\\
Matthew Watson\\
Meg Risdal\\
Mehran Kazemi\\
Michael Moynihan\\
Ming Zhang\\
Minsuk Kahng\\
Minwoo Park\\
Mofi Rahman\\
Mohit Khatwani\\
Natalie Dao\\
Nenshad Bardoliwalla\\
Nesh Devanathan\\
Neta Dumai\\
Nilay Chauhan\\
Oscar Wahltinez\\
Pankil Botarda\\
Parker Barnes \\
Paul Barham\\
Paul Michel\\
Pengchong Jin\\
Petko Georgiev\\
Phil Culliton\\
Pradeep Kuppala\\
Ramona Comanescu\\
Ramona Merhej\\
Reena Jana\\
Reza Ardeshir Rokni\\
Rishabh Agarwal\\
Ryan Mullins\\
Samaneh Saadat\\
Sara Mc Carthy\\
Sarah Cogan \\
Sarah Perrin\\
Sébastien M. R. Arnold\\
Sebastian Krause\\
Shengyang Dai\\
Shruti Garg\\
Shruti Sheth\\
Sue Ronstrom\\
Susan Chan\\
Timothy Jordan\\
Ting Yu\\
Tom Eccles\\
Tom Hennigan\\
Tomas Kocisky\\
Tulsee Doshi\\ 
Vihan Jain\\
Vikas Yadav\\
Vilobh Meshram\\
Vishal Dharmadhikari\\
Warren Barkley\\
Wei Wei\\
Wenming Ye\\
Woohyun Han\\
Woosuk Kwon\\
Xiang Xu\\
Zhe Shen\\
Zhitao Gong\\
Zichuan Wei\\
\\
\noindent\textbf{Support}\\
Victor Cotruta\\
Phoebe Kirk\\
Anand Rao\\
Minh Giang\\
Ludovic Peran\\
Tris Warkentin\\
\\
\noindent\textbf{Sponsors}\\
Eli Collins\\
Joelle Barral\\
Zoubin Ghahramani\\
Raia Hadsell\\
D. Sculley\\
Jeanine Banks\\
Anca Dragan\\
Slav Petrov\\
Oriol Vinyals\\
Jeff Dean\\
Demis Hassabis\\
Koray Kavukcuoglu\\
Clement Farabet\\
\\
\noindent\textbf{Technical advisors}\\
Elena Buchatskaya\\
Sebastian Borgeaud\\
Noah Fiedel \\
\\
\noindent\textbf{Lead}\\
Armand Joulin\\
\\
\noindent\textbf{Technical leads}\\
Kathleen Kenealy\\
Robert Dadashi\\
Alek Andreev
\clearpage

\bibliography{main}

\clearpage
\appendix
\onecolumn

\end{document}